\documentclass[]{iscram}

\usepackage[utf8]{inputenc}

\usepackage{lipsum}
\usepackage{tikz}
\usepackage{fancyvrb}
\usepackage{listings}
\usepackage{enumitem}
\usepackage{tcolorbox}
\usepackage{multicol}
\usepackage{siunitx}
\lstdefinestyle{common}{
  xleftmargin=.5em,
  xrightmargin=.5em,
  frame=single,framesep=.5em,framerule=0pt,
  fancyvrb=true,
  basicstyle=\ttfamily,
  keywordstyle=\color{cyan!50!blue!75!black}\bfseries,
  commentstyle=\color{red!50!black}\itshape,
  stringstyle=\ttfamily\color{green!50!black},
  numbers=none,
  showspaces=false,
  showstringspaces=false,
  fontadjust=true,
  keepspaces=true,
  flexiblecolumns=true,
  emphstyle=\color{red},
}
\lstdefinestyle{TeX}{
  style=common,
  backgroundcolor=\color{blue!5},
  aboveskip=5pt,
  belowskip=5pt,
  language=[LaTeX]TeX,
  moretexcs={
    abstract, addbibresource, iscramset, keywords, mainmatter,
    maketitle, printbibliography, subsection, subsubsection, url,
    urldef, href, includegraphics, ldots, parencite, citeauthor,
    citeyear, citetitle, midrule, toprule, bottomrule
  },
  fancyvrb=true,
}
\lstdefinestyle{console}{
  style=common,
  backgroundcolor=\color{gray!10},
  aboveskip=5pt,
  belowskip=5pt,
}

\newlist{options}{description}{1}
\setlist[options]{%
  beginpenalty=10000,%
  itemsep=.5\parskip plus .3\parskip minus .2\parskip,
  parsep=.5\parskip plus .3\parskip minus .2\parskip,
  topsep=.5\parskip plus .3\parskip minus .2\parskip,
  partopsep=.5\parskip plus .3\parskip minus .2\parskip,
  style=nextline,labelindent=1em,%
  font=\normalfont\ttfamily}

\colorlet{macro color}{cyan!50!blue!75!black}
\colorlet{option color}{red!50!black}
\colorlet{generic color}{green!40!black}

\newtcolorbox{pseudoTeX}{colback=blue!5,colframe=blue!5,before=\nobreak}
\let\LaTeXorig\LaTeX
\renewcommand\LaTeX{\bgroup\fontfamily{lmr}\selectfont\upshape\LaTeXorig\egroup}

\iscramset{
  title={
    How does a Pre-Trained Transformer Integrate Contextual Keywords? Application to Humanitarian Computing
  },
  short title={How does a Pre-Trained Transformer Integrate Contextual Keywords?
  },
 author={
   short name={Valentin Barriere},
   full name=Valentin Barriere\thanks{corresponding author},
   affiliation={
     European Commission \\ Joint Research Centre (JRC) - Ispra %
       \\
     \href{mailto:valentin.barriere@ec.europa.eu}{\url{valentin.barriere@ec.europa.eu}}
   },
 },
 author={
   full name=Guillaume Jacquet,
   affiliation={
     European Commission \\ Joint Research Centre (JRC) - Ispra \\
     \href{mailto:guillaume.jacquet@ec.europa.eu}{\url{guillaume.jacquet@ec.europa.eu}}
   },
 },
}
\usepackage{tikzpagenodes}
\usetikzlibrary{fit}

\usepackage{times}
\usepackage{latexsym}

\usepackage[T1]{fontenc}

\usepackage[utf8]{inputenc}

\usepackage{microtype}

\usepackage{multirow}
\usepackage{tcolorbox}

\usepackage{tabularx} 
\usepackage{graphicx, wrapfig, subcaption, setspace, booktabs}
\usepackage{url}
\usepackage{latexsym}
\usepackage{graphicx}
\usepackage{pifont}

\newcommand{\cmark}{\ding{51}}%
\newcommand{\xmark}{\ding{55}}%

\usepackage{adjustbox}

\usepackage{flushend}

\usepackage{verbatim}

\newcommand{\rougeval}[1]{\textcolor{red}{\textbf{{#1}}}}
\newcommand{\vertval}[1]{\textcolor{olive}{\textbf{{#1}}}}

\usepackage{courier}

\usepackage{xcolor}
\newcommand{\corrval}[2]{{#2}}
\newcommand{\addval}[1]{{#1}}

\begin{document}
	
	\maketitle
	
	\makeatletter
	\makeatother
	
	\abstract{
		In a classification task, dealing with text snippets and metadata usually requires to deal with multimodal approaches. When those metadata are textual, 
		it is tempting to use them intrinsically with a pre-trained transformer, in order to leverage the semantic information encoded inside the model. 
		This paper describes how to improve a humanitarian classification task by adding the crisis event type to each tweet to be classified. 
		Based on additional experiments of the model weights and behavior, it identifies how the proposed neural network approach is partially over-fitting the particularities of the Crisis Benchmark, to better highlight how the model is still undoubtedly learning to use and take advantage of the metadata's textual semantics.
	}
	
	
	
	\section{Introduction}
	
	It is frequent to fuse information from different channels in order to help the classifier disambiguate some examples \cite{Xu2013}. 
	Nonetheless, systems that mix together the different information fluxes inside the model are mainly used in multimodal data-processing \cite{Baltrusaitis} and less frequent for a simple modality. 
	
	The rise of Transformer models \cite{Vaswani2017} created a new set of possibilities in text processing.
	Nowadays, the new architectures are pushing the boundaries of the state-of-the-art \cite{Raffel2019,Brown2020} in general tasks and no longer only in Machine Translation. 
	
	Those architectures allow us to efficiently process an input composed of 2 sentences, like for Natural Language Inference tasks \cite{Williams2017}, so that they can interact together and the fusion happens inside the model. 
	This is interesting since the model is able to infer the meanings contained in both sentences in context.
	
	In this paper, we propose the use of textual metadata semantics, by encoding it textually in the input of the classifier
	\corrval{. In this way,}{ in a way} the model can detect metadata as separate information and yet is able to use cross-input attention between the metadata and the text content. 
	This can be seen as related to some works done on zero-shot learning using Natural Language Inference (NLI) where the semantic content of the label is encoded inside the classifier \cite{Yin2020,Halder2020}. 
	\addval{\cite{Clark2019} studied the attention mechanism of a BERT model and clustered the attention heads. \cite{Rogers2020a} reviewed the understanding of BERT-based transformers in the literature, highlighting that semantics information was spread across the entire model
		. 
		From the best of our knowledge, no work has been done to analyze semantics between a contextual keyword and its associated sentence by retrieving the most important words with the help of the attention weights, then clustering the retrieved words embeddings.}

	In the context of a crisis caused by a natural or human source, it is important to be react quickly and to be aware of the population needs \cite{Alam2020}. The resources created by social media users can contain crucial information \cite{Olteanu2015, Imran2016b} because of the poster witnessing the situation directly \cite{Zahra2020}.
	The type of event is highly variable, and the system needs to be relevant in the context of a natural disaster like a fire or a hurricane \cite{Waqas2019, Alam2018a}, a train crash or a pandemic (\cite{Qazi2020}). Moreover, it is known that out-of-domain prediction is harder \cite{Alam2018b}. 
	
	We chose to work on the Crisis Benchmark 14-event-types dataset \cite{Alam2021benchmark} on a 11-humanitarian-class supervised setting, in order to use the event type as metadata to enhance the quality of the predictions, comparing the results using BERT \cite{Devlin2018}, RoBERTa \cite{Liu2019} and T5 \cite{Raffel2019}.\addval{ It is important to note that in this type of task, the urge of retrieving information from social media arrives after the beginning of the disaster, hence, the type of disaster is always known and it is an information available for use.} 
	We found that in every configuration the event-aware models obtain better results than their Vanilla counterparts. We then investigated this improvement by an intensive study of the dataset, the learned model behaviors and predictions. In addition to the experiments using the official partition, we used a Leave-One-Event-Type-Out setting (LOETO) in order to prevent the model from overfitting some particularities of the dataset events. \addval{This work is directly in continuation with the work of \cite{Alam2021benchmark}, in which they add the event type to the tweet content using a simple concatenation, leading to no improvement over the Vanilla model.}
	
	%
	%
	%
	%
	%
	%
	%
	%
	%
	%
	%
	%
	
	\vspace*{-.1cm}
	\section{Method}
	\vspace*{-.1cm}
	The proposed approach consists of 
	the integration of metadata information inside the input text representation. We chose a dataset where this was available and where the authors of the original dataset claimed that 
	their method 
	was not working. We decided to integrate the event type into the learning algorithm using a NLI configuration for 3 transformer models. 
	After running several experiments, we observed an increase in all the models' performances, independently of the 
	architecture. 
	We ran some analyses in order to investigate whether this gain came from the textual content of the metadata, or from mechanical memorization of the label distribution conditioned by the metadata. 
	
	
	We chose to focus on the Crisis Benchmark's humanitarian classification task for space reasons. The reader should note that we also made experiments on the informativeness binary classification task described in \citet{Alam2021benchmark} that lead to positive results. 
	
	
	
	\vspace*{-.1cm}
	\subsection{Incorporating Event Type} \vspace*{-.1cm}
	
	Instead of naively concatenating the event-type and the tweet together, 
	we used the patterns offered by the different models to separate the two types of information. In this way, we are not breaking the syntax and the rhythm of the tweet sentence by adding a piece of text that does not belong in the initial sentence. 
	For BERT and RoBERTa, we used the special tokens in order to separate the 2 sentences. 
	For T5, we used a text pattern, with a new task name and the \texttt{sentence} and \texttt{context}. Examples of the different pre-processing configurations we used are visible in Table \ref{tab:config_preproc}.

	\begin{table*}[]
		\centering
		\caption{Examples of text pre-processing for each model}
		\resizebox{.9\textwidth}{!}{%
			\begin{tabular}{l|l}
				Model   & Example     \\ \hline
				BERT    & \footnotesize{\texttt{{[}CLS{]} fire {[}SEP{]} After deadly Brazil nightclub fire, safety questions emerge. {[}SEP{]}}}                                                  \\
				RoBERTa & \footnotesize{\texttt{\textless s\textgreater fire \textless/s\textgreater After deadly Brazil nightclub fire, safety questions emerge. \textless/s\textgreater}} \\
				T5     & \footnotesize{\texttt{cbmk context: fire sentence: After deadly Brazil nightclub fire, safety questions emerge.}}                                                     
			\end{tabular}
		}
		\label{tab:config_preproc}
		\vspace*{-.5cm}
	\end{table*}
	
	By calling $y$ the humanitarian label, $\mathbf{x}$ the observed tweet and $m$ the metadata, we can reformulate the prediction of the Vanilla model as $P(y|\mathbf{x})$. Instead of modeling $P(y|\mathbf{x})$, the event-aware configuration models a conditioned probability $P(y|\mathbf{x}, m)$. Another way to integrate the metadata would be to model the joint probability of the class $y$ and the metadata $m$ in $P(y, m|\mathbf{x})$. We chose the conditioned model in order to easily use the semantic information contained inside the keyword, similarly to the way some zero-shot NLI models are handling the labels \cite{Yin2020}. 
	
	
	\vspace*{-.1cm}
	\subsection{Model and dataset analysis} \vspace*{-.1cm}
	
	We ran some analysis of the event-aware model, in order to verify if the model was integrating semantic information regarding the textual content of the event. Transformers and more generally Neural Networks models are known to be black boxes, but some works have been done in the direction of interpreting them \cite{Clark2019}. In combination with analysis of the model and its behavior, we also analyze the dataset and its label distribution, allowing us to detect some patterns on which the neural network could rely on. 
	
	We seek to answer the following questions:
	
	
	\begin{tcolorbox}
		\textbf{-} \textit{Dataset label distribution}: What does the labels distribution look like for each event ? 
		
		\textbf{-} 
		\textit{Predicted label distribution}: What is the impact of conditioning over an event on the predictions distribution?
		
		\textbf{-} \textit{Out-of-domain learning}: Is the event-aware model still better on a Leave-One-Event-Type-Out setting? 
		
		\textbf{-} \textit{Attention weights}: What words are influenced by the metadata event type token? 
		
	\end{tcolorbox}
	
	We ran the models with the \texttt{transformers} library \cite{Wolf2019}. 
	For BERT and RoBERTa, we used a learning rate of 1e-6, Adam algorithm, for maximum 20 epochs, while we ran the T5 for 3 epochs, using a learning rate of 3e-4.  
	For RoBERTa, we manually added and trained a new layer\footnote{Layer managing the \texttt{token\_type\_ids}} allowing to specify the token types, which is normally not used during RoBERTa pre-training. 
	On the LOETO setting, we created the dev by taking 25\% of the test set, otherwise we used the official train/dev/test partition. 
	We normalized the size of every sequence up to 128 tokens in total. 
	
	\vspace*{-.2cm}
	\section{Experiments and Analysis} \vspace*{-.1cm}
	
	\subsection{Dataset} \vspace*{-.1cm}
	
	For our experiments, we used the Crisis Benchmark dataset \cite{Alam2021benchmark} composed of 87,557 tweets collected during several crisis events, that can be separated into 14 event types: bombing, collapse, crash, disease, earthquake, explosion, fire, flood, hazard, hurricane, landslide, shooting, volcano, or none. 
	This dataset has been labeled into 11 humanitarian classes: \textit{Affected individuals, Caution and advice, Displaced and evacuations, Donation and volunteering, Infrastructure and utilities damage, Injured or dead people, Missing and found people, Not humanitarian, Requests or needs, Response efforts, Sympathy and support}. 
	We refer the readers to the original paper for more details on the dataset \cite{Alam2021benchmark}. 
	
	\vspace*{-.2cm}
	\subsection{Classifier results} 
	\vspace*{-.1cm}
	
	The results of the experiments comparing the event-aware and event-unaware models are shown in Table \ref{tab:results}. We used unweighted means of the Precision, Recall and F1, as well as global Accuracy and weighted F1. The best results among the transformers are obtained without surprise with the T5, which has 220M parameters, more than BERT and RoBERTa (110 and 125M). 
	For each transformer, the event-aware model reaches higher performances compared to Vanilla model.
	The results event per event are available in Table \ref{tab:per_event}, the only event that is worse in the event-aware setting is '\textit{fire}'. 
	Finally, we believe the reason our Vanilla models obtain significantly better results than \citet{Alam2021benchmark} remains in a better fine-tuning of the transformers, with a lower learning rate and a higher number of epochs. 
	
	\begin{table}[]
		\centering
		\caption{Results on the humanitarian classification task. 
		}
		\begin{tabular}{c|c|ccc|cc}
			Model & Event & Prec & Rec & u-F1  & w-F1 & Acc \\ \hline \hline
			BERT \cite{Alam2021benchmark} &   \cmark   &    70.1 &   71.3  &  70.7 & 86.5 & 86.5    \\ 
			RoBERTa \cite{Alam2021benchmark}&   \cmark   &    70.2  &   72.3  &  71.1 & 87.0 & 87.0    \\ \hline
			\multirow{2}{*}{BERT} &   \xmark   & 73.5    &   71.9  &  72.5  & 87.5 & 87.5    \\
			&     \cmark        &  75.3   &   72.5  & 73.7   & 88.3  & 88.1    \\ \hline
			\multirow{2}{*}{RoBERTa}      &    \xmark    &  74.2    &   73.6  &  73.7  & 87.9 & 88.0   \\
			&        \cmark      &  74.1  &    74.5 &  74.1  &   88.5 &   88.5  \\ \hline
			\multirow{2}{*}{T5} &  \xmark   &   75.0   &  74.4   &  74.6  & 88.3 & 88.4   \\
			&        \cmark      &  76.7   &  73.8   &   \textbf{75.1} & \textbf{88.8} &   \textbf{88.9}  
		\end{tabular}
		\label{tab:results}
	\end{table}

	\vspace*{-.2cm}
	\subsection{Analysis} 
	\vspace*{-.1cm}
	We run analyses on the dataset as well as BERT weights and behavior in order to interpret the good results of the event-aware model compared to the Vanilla model. 
	
	\vspace*{-.1cm}
	\subsubsection*{Dataset label distribution} 
	
\begin{figure*}
		\centering
		\hspace*{-1.cm}
		\includegraphics[width=1.15\textwidth]{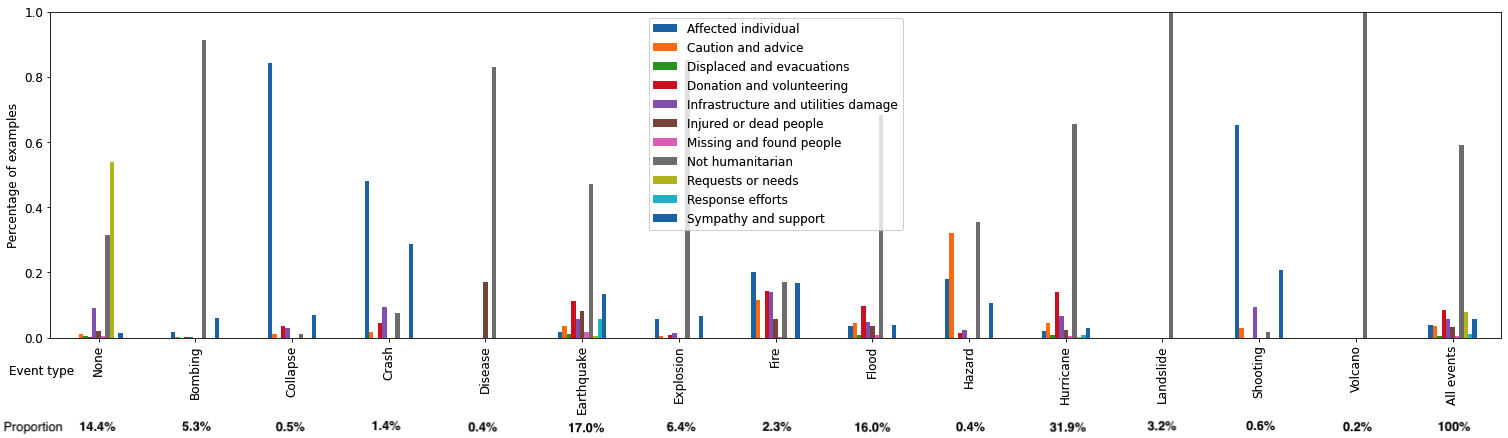}
		\caption{Distributions of labels regarding the event type in the train set, with the proportion of each event type}
		\label{fig:dist_labels_per_event}
	\end{figure*}

	It is interesting to look at the distribution of the labels regarding the event type, available in Figure \ref{fig:dist_labels_per_event} for the training set\footnote{The distributions are comparable on the test set}. We can see the type of labels appearing in each event type is highly imbalanced: for some events, like \textit{Landslide} and \textit{Volcano}, there is even only one label, which is \textit{Not humanitarian}. These events, which are highly unrepresentative in term of labels distribution, only represent a small portion of the dataset (Figure \ref{fig:dist_labels_per_event}). 
	
	By comparing the distributions with the model's performances, we can see that the only event type where the event-aware is worst is '\textit{fire}', where the distribution of labels is closer to uniform. 
	
	
	With such diverse distributions of labels regarding the event, it is obvious that memorizing the label distribution of each event will be highly beneficial for the model, but just an overfitting of the model on the dataset.  
	
	\subsubsection*{Predicted Label Distribution} 
	
	In order to check if conditioning over an event was distorting the predictions, 
	we calculated the Kullback-Leibler divergences between the test set distribution $\mathcal{D}_{t}$, the distributions of predictions over the test set conditioned by an event $E$, $\mathcal{D}_{t}(E)$, and the event's label distributions $\mathcal{D}_{e}(E)$. We found that, when conditioned over an event, the model predicts on average over the test set, a distribution that is closer to its respective event distribution (0.62) than to the test set distribution (0.69) (Eq. \ref{eq:kl}). 
	This seems to confirm the impact of the event type token in overfitting some dataset particularities.
	
	\begin{equation}
	\sum_{E}KL(\mathcal{D}_{e}(E)||\mathcal{D}_{t}(E)) < \sum_{E}KL(\mathcal{D}_{t}||\mathcal{D}_{t}(E))
	\label{eq:kl}
	\end{equation}
	\normalsize

	\subsubsection*{Leave One Event Type Out} 
	
	We ran a LOETO 
	cross-validation in addition to the experiments using the official partition of the dataset.
	This setting allows the comparison of results in a situation where the event-aware model is not able to infer the label distribution of the event type. 
	
	
	This method obtains necessarily worse results than when testing on samples from an event type that was used during learning, the goal is to compare the Vanilla BERT and the event-aware BERT. If the event-aware model reaches higher performances than the Vanilla, then adding textual metadata has a positive impact on the performances. 
	
	\begin{table}[]
		\centering
		\caption{Results of the BERT model on LOETO}
		\begin{tabular}{l|lll|l}
			Model type& Prec & Rec & F1 & Acc \\ \hline \hline
			Vanilla     &  40.0    & 54.9    & 44.1   &  65.4   \\ 
			Event-aware &   47.0   & 55.2    &  45.2  &   \textbf{67.6}  \\
		\end{tabular}
		\label{tab:loete}
	\end{table}

	\begin{table}[]
		\centering
		\caption{Accuracies (differences with Vanilla) for each event type of the event-aware BERT on the humanitarian classification task, for official partition and LOETO}
		\begin{tabular}{llllllll}
			\multicolumn{1}{l|}{Partition}  & None & Bombing & Collapse & Crash & Disease & Earthquake & Explosion \\ \hline \hline
			\multicolumn{1}{l|}{Official} &   91.2 (\vertval{1.2})   &   96.7 (\vertval{0.4})      &    88.8 (0.0)      &   89.3 (\vertval{1.1})  & 98.6 (\vertval{2.9})     &    77.0 (\vertval{1.2})   &       96.6 (\vertval{0.3})     \\
			\multicolumn{1}{l|}{LOETO}  &  34.3 (\vertval{5.0})    &    89.7 (\rougeval{-4.3})    &     44.1 (\vertval{19.7})     &      81.5 (\rougeval{-0.3}) & 59.4 (\rougeval{-11.3})    &   49.4 (\rougeval{-1.6})   &   93.1 (\vertval{1.4})   \\
			&  &  &  & & & &  \\
			 Fire & Flood & Hazard  & Hurricane & Lanslide & Shooting & Volcano & \textbf{Average}  \\ \hline \hline
			   81.5 (\rougeval{-1.2})  &    90.7 (\vertval{0.7}) & 52.8 (0.0)    &   88.0   (\vertval{0.6})      &  100  (\vertval{1.6})      &      87.5 (0.0)     &      97.1 (0.0)  &  88.3 (\vertval{0.8})  \\
			 67.6 (\rougeval{-4.2})     &  85.3 (\vertval{1.7})  & 49.8 (\vertval{1.4})   &   71.7 (\vertval{5.0})   &    92.6 (\rougeval{-0.6})       &  77.8 (\vertval{7.1})       &      72.0 (\rougeval{-2.8})  & 67.6 (\vertval{2.2})  \\

		\end{tabular}
		\label{tab:per_event}
	\end{table}

	
	The results shown in Table \ref{tab:loete} are explicit: adding the event type also improves the results when the system is faced with a new event it had not seen before. In Table \ref{tab:per_event} we can see on the results that the event-aware model is not homogeneously performing according to event types. 

	\subsubsection*{Attention weights} 
	
	In order to understand how the event type is influencing the decision, we look at the interaction mechanism between the event type and the tweet by studying the attention weights of the BERT model. 
	To avoid any model overfitting on specific words corresponding to specific events, we ran this study in a LOETO configuration, so that the model has never seen the event type token before and cannot make a correlation between this token and words appearing for this event type. 
	
	For each event, we counted the number of times every token from the tweet was linked to the event token with an attention weight bigger than an arbitrary threshold of 0.5. 
	We discarded the weights between 
	the punctuation and the stop-words\footnote{from nltk toolbox} tokens. Then we took the 50 tokens with the highest tf-idf, extracted their embeddings with a BERT model, and proceeded with a clustering. 
	
\begin{figure}
		\centering
		\includegraphics[width=.8\textwidth]{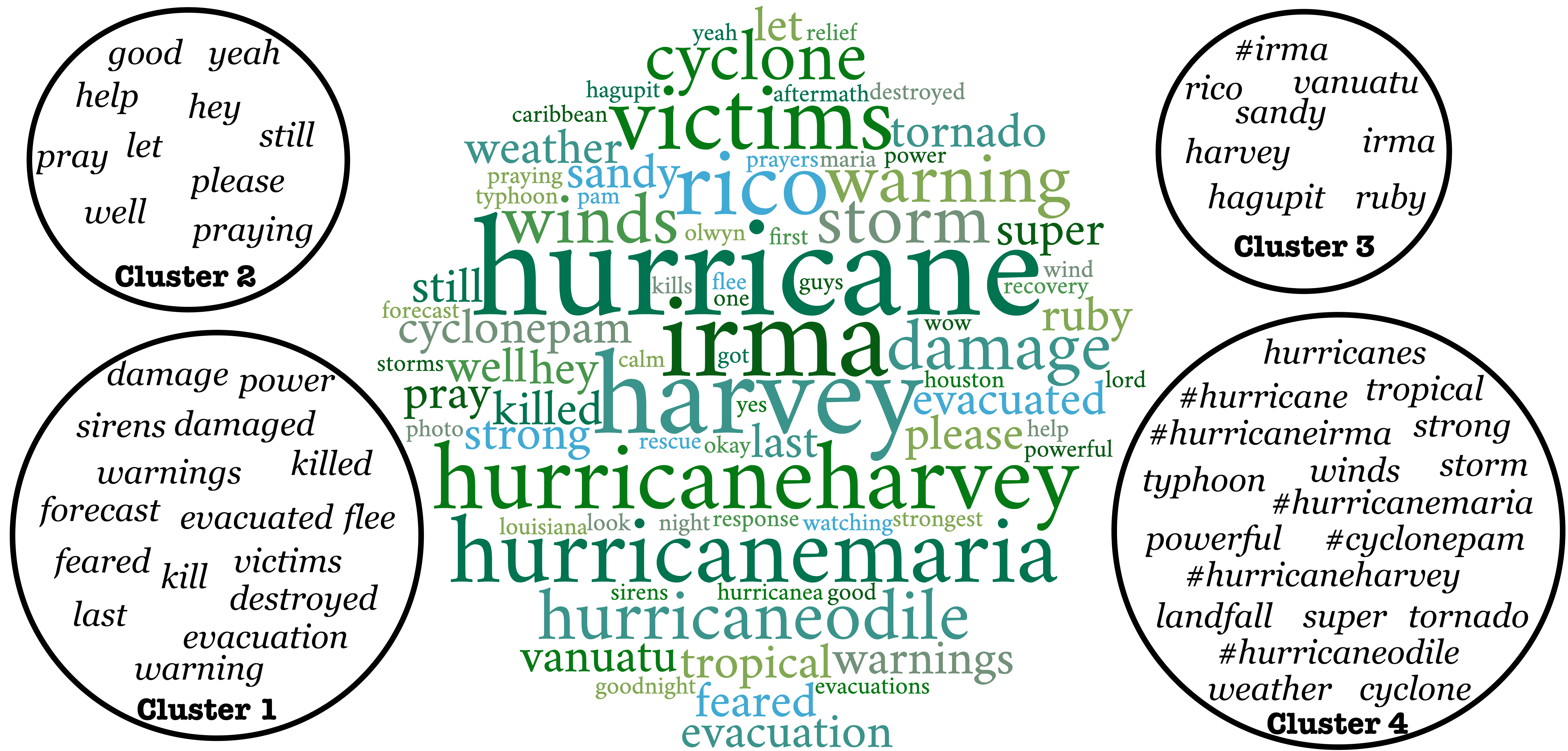}
		\caption{Tokens interacting the most with the event type '\textit{hurricane}'. Clusters of the top-50 tokens.}
		\label{fig:cloud}
	\end{figure}

	A visualization of the words and clusters for \textit{hurricane} has been made in Figure \ref{fig:cloud}
	. 
	We figured out that the tokens interacting directly with the event type were related to the type of disaster\footnote{\textit{hurricane}, \textit{cyclone}, \textit{storm}, \textit{tornado}, \textit{typhoon}...}, proper names\footnote{\textit{Irma}, \textit{Sandy}, \textit{Harvey}, \textit{Vanuatu},…}, and the classes of the task.\footnote{sympathy, material damages, human damages, warnings, evacuations,...} 
	This means that even for an event type not contained in the training set, the model is capable of using the semantics of this event in order to better infer the class of the tweet. 
	
	%
	%
	
	\section{Conclusion} 
	
	In this article, we studied the effect of adding textual metadata information inside three transformer models, using the classical ways to separate different sentences in an input. 
	
	We ran experiments on a Humanitarian Computing dataset, adding to each tweet its respective event type\addval{, which is contextual information always available,} in order to better classify it into 11 classes. 
	We discovered that this method improves the results,\addval{ even when the event type has never been seen during training phase. It} can be applied to different transformers and we obtained a new state-of-the-art on this dataset. Finally, we carried out an analysis of the dataset and the event-aware BERT model weights and behavior in order to shed light on the reasons for this increase in performance. 
	
	We show that the event-aware model is not only memorizing the unbalanced label distribution of the event type, but also learning semantics relation between the text and the event type token. 
	We discovered that the tokens locally interacting with the event token were related to the lexical fields of the event type, and the classes of the classification task.
	
	\newpage
	
	\bibliography{JRC}
	\bibliographystyle{acl_natbib}
	
\end{document}